\documentclass[10pt, fleqn]{article}
\usepackage[utf8]{inputenc}
\DeclareUnicodeCharacter{2212}{$-$}
\usepackage{graphicx}
\usepackage[fleqn]{amsmath}
\usepackage[version=4]{mhchem}
\usepackage{siunitx}
\usepackage{longtable,tabularx}
\usepackage{nccmath}
\usepackage{amsmath}
\usepackage{algorithm}
\usepackage{algpseudocode}
\usepackage[superscript]{cite}
\usepackage[margin=1in]{geometry}
\usepackage[toc,page]{appendix}
\usepackage{setspace}
\usepackage{amssymb}
\usepackage{tabu}
\usepackage{listings}
\usepackage{afterpage}
\usepackage{titlesec}
\usepackage[labelfont=bf, font=small,skip=4pt]{caption}
\usepackage{fancyhdr}
\usepackage{pgf}
\usepackage{subcaption}
\usepackage{textcomp}
\usepackage{pict2e}
\usepackage{wasysym}
\usepackage{multicol}
\usepackage{float}

\usepackage{lipsum}
\usepackage{physics}
\usepackage{booktabs}
\usepackage{enumitem}
\usepackage{indentfirst}
\usepackage{times}
\usepackage{cite}
\usepackage{multirow}
\usepackage{makecell}
\usepackage{hyperref}
\hypersetup{
    colorlinks=true,
    linkcolor=blue,
    filecolor=magenta,      
    urlcolor=cyan,
    }
\usepackage{cleveref}
\usepackage{amsmath,amssymb,amsfonts}
\usepackage{mathtools}
\usepackage{comment}
\usepackage{textcomp}
\usepackage{xcolor}
\usepackage{float}
\usepackage{soul}
\usepackage{enumitem}
\usepackage{cancel}
\usepackage{algorithm}
\usepackage{algpseudocode}
\usepackage{amsthm}
\usepackage{amsmath}
\usepackage{wrapfig}
\usepackage{color, colortbl}
\usepackage{stackengine}

\definecolor{LightCyan}{rgb}{0.4, 0.9, 0.8}

\DeclarePairedDelimiter{\floor}{\lfloor}{\rfloor}

\definecolor{Melon}{rgb}{1.0, 0.8, 0.4}

\usepackage{color, colortbl}

\titleformat{\section}
{\normalfont\normalsize\bfseries}{\thesection}{0.5em}{}
\titleformat{\subsection}
{\normalfont\normalsize\bfseries\em}{\thesubsection}{0.5em}{}
\titleformat{\subsubsection}
{\normalfont\normalsize\em}{\thesubsubsection}{0.5em}{}
\begin{document}

\begin{flushright}
\Large %

\textbf{SSC24-P3-15}
\end{flushright}
\begin{centering}      
\large %

\textbf{Vision-Based Detection of Uncooperative Targets and Components on Small Satellites}\\
\vspace{0.5cm}
\normalsize %

{Hannah Grauer$^*$}, Elena-Sorina Lupu$^*$, Connor Lee, and Soon-Jo Chung\\
{California Institute of Technology}\\
{1200 E. California Blvd., Pasadena, CA, 91125}\\
{\texttt{\{hgrauer, eslupu, clee, sjchung\}}@caltech.edu} \\ 
\vspace{0.5cm}
Darren Rowen, Benjamen Bycroft, Phaedrus Leeds, and John Brader\\
{The Aerospace Corporation}\\
{2310 E. El Segundo Blvd., El Segundo, CA 90245-4609}\\
{\texttt{\{darren.w.rowen, benjamen.p.bycroft, phaedrus.leeds, john.s.brader\}@aero.org}}\\
\vspace{0.5cm}
$^*$ These authors contributed equally

\vspace{0.5cm}
\centerline{\textbf{ABSTRACT}}
\vspace{0.3cm}
\end{centering}

Space debris and inactive satellites pose a threat to the safety and integrity of operational spacecraft and motivate the need for space situational awareness techniques. These uncooperative targets create a challenging tracking and detection problem due to a lack of prior knowledge of their features, trajectories, or even existence. Recent advancements in computer vision models can be used to improve upon existing methods for tracking such uncooperative targets to make them more robust and reliable to the wide-ranging nature of the target. This paper introduces an autonomous detection model designed to identify and monitor these objects using learning and computer vision. The autonomous detection method aims to identify and accurately track the uncooperative targets in varied circumstances, including different camera spectral sensitivities, lighting, and backgrounds. Our method adapts to the relative distance between the observing spacecraft and the target, and different detection strategies are adjusted based on distance. At larger distances, we utilize You Only Look Once (YOLOv8), a multitask Convolutional Neural Network (CNN), for zero-shot and domain-specific single-shot real time detection of the target. At shorter distances, we use knowledge distillation to combine visual foundation models with a lightweight fast segmentation CNN (Fast-SCNN) to segment the spacecraft components with low storage requirements and fast inference times, and to enable weight updates from earth and possible onboard training. Lastly, we test our method on a custom dataset simulating the unique conditions encountered in space, as well as a publicly-available dataset.

\begin{multicols*}{2}

\section{INTRODUCTION}

On-orbit services (OOS)~\cite{aerospace} is a growing sector of the space industry that involve various activities aimed at extending the life, enhancing the capabilities, or re-purposing spacecraft and other objects in space.
Some examples of OOS are in-space servicing and inspection~\cite{doi:10.2514/6.2016-5478}, rendezvous and docking~\cite{FOUST2020191, 9061445}, or debris removal.~\cite{MARK2019194, cleanspaceone, Fehse_2003, NISHIDA200995}
To make these services viable for small satellites (SmallSats), we must develop advanced autonomous capabilities that reduce costs and minimize the need for human intervention.
As an example, in-space servicing and inspection enable the repair and upgrade of satellites while in orbit.
One of the key benefits of these technologies is cost reduction. 
By repairing defunct spacecraft in orbit, the high costs associated with launching new spacecraft can be reduced.

The procedure for in-space servicing involves a sequence of steps such as intercepting with the spacecraft, close-proximity operations and eventually rendezvous and docking.
Oftentimes, the servicing vehicle deployed in-orbit does not need to be a large spacecraft.
Instead, a SmallSat equipped with necessary sensors, compute, and autonomy is more advantageous from a cost and weight standpoint. 
However, as their size is further decreased (i.e., nanosatellites and cubesats) the sensors and compute tend to be limited \cite{bandyopadhyay2016review}.

To achieve autonomous OOS, it is important to first detect and continuously track the target spacecraft from afar. 
As the servicing spacecraft approaches the target, it must be equipped to accurately identify specific components for maintenance or find suitable attachment points to facilitate the de-orbiting of the spacecraft using on-board cameras.
On the ground, computer vision models using machine learning have made significant progress in object detection~\cite{object_detection_review} and segmentation~\cite{semantic_segmentation, awais2023foundational}. However, these achievements have yet to be fully applied in space. Like field robotic perception development in terrestrial applications~\cite{lee2024cart, gan2023unsupervised, lee2023online}, this shortfall is primarily due to the limited availability and small sizes of publicly-available datasets available for training, resulting in less robust model performance. Additionally, the computational resources required for both training and inference in space pose a significant challenge in terms of on-board compute, on-board storage, and Earth-to-satellite latency.

\subsection*{Contributions}
\begin{figure*}[htbp]
    \centering
    \includegraphics[width=\linewidth]{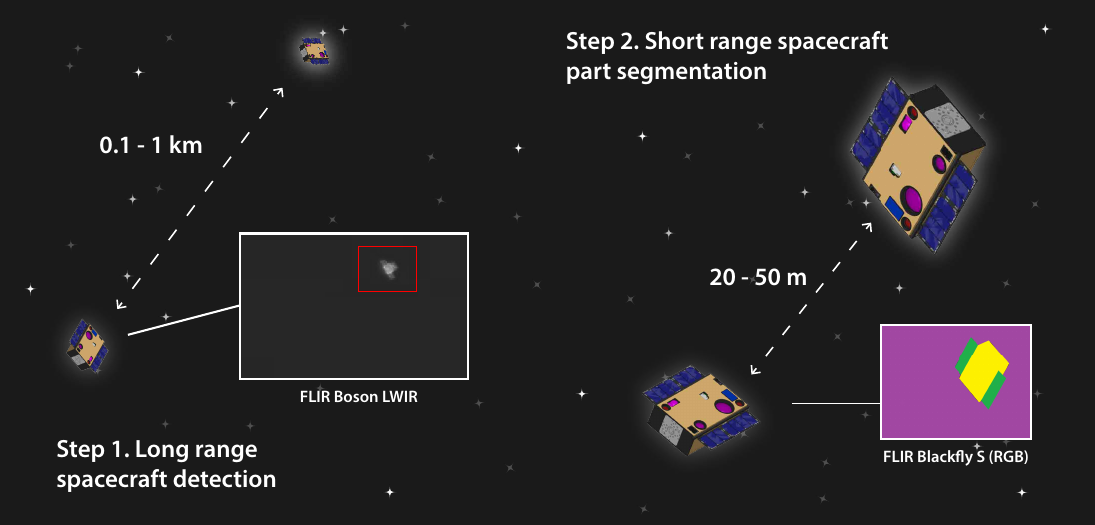}
    \captionof{figure}{Overview of the long-short detection architecture proposed for the Edge Node mission or other OOS tasks.~\cite{EdgeNode} We propose a long distance detection model on thermal images for detection at larger ranges into  a segmentation model that identifies spacecraft parts for OOS on RGB images.}
    \label{fig:flowchart}
\end{figure*}

To address these limitations we present a long and short range methodology that is able to track a spacecraft from afar and identify spacecraft components using segmentation, as demonstrated by Figure \ref{fig:flowchart}.
Our contributions are as follows:
\begin{itemize}[noitemsep, leftmargin=*]
\item A novel method to generate long-range images for training using thermal cameras, which are more robust to varied lighting conditions in space.
\item A long range detection algorithm operating on thermal images of a spacecraft using the YOLOv8 (You Only Look Once) model~\cite{ultralytics_yolo_2023}.
\item A novel method for spacecraft part segmentation at a shorter range that leverages the performance of a visual foundation model (VFM) distilled into a lightweight fast semantic network to increase segmentation accuracy while keeping inference and training time low.
\end{itemize}

Our methodology is designed to align with Aerospace Corporation's Edge Node and risk-reduction Edge Node Lite missions for formation flight and rendezvous and proximity operations.
These missions use multiple small satellites equipped with miniature sensor and computing hardware.
We use the Edge Node spacecraft as the target. 
We additionally use from the missions the FLIR Boson+ Long-wave infrared (LWIR) camera in our custom dataset creation and the Jetson TX2 NX to test performance.~\cite{EdgeNode}

\section{RELATED WORK}

\subsection{Long Range Object Detection}

In the domain of spacecraft pose estimation, recent works have adopted various computer vision models, particularly focusing on the use of object detectors based on convolutional neural networks (CNNs). While current state-of-the-art computer vision benchmarks, including object detection, are dominated by transformer-based architectures~\cite{dosovitskiy2020image, liu2021swin, carion2020end}, CNNs exhibit higher computational efficiency on embedded devices and are more suited for real-time, low-compute robotic applications. In field robotic applications where publicly-available curated datasets are rare or limited, CNNs are also more favorable due to their built-in spatial inductive bias (via 2D convolutions) that makes them less prone to overfitting when training with smaller datasets~\cite{dosovitskiy2020image, liu2022convnet, lee2024cart, lee2023online, gan2023unsupervised}. 

Current CNN-based object detectors fall into one of two categories: two-stage and one-stage detectors. Two-stage detectors include models like Faster-RCNN~\cite{ren2015faster} which perform initial region proposal steps before moving forward to bounding box refinement and classification. On the other hand, single-stage detectors like SSD~\cite{liu2016ssd}, YOLOv3~\cite{redmon2018yolov3}, RetinaNet~\cite{lin2017focal}, and EfficientDet~\cite{tan2020efficientdet} perform bounding box refinement and classification in one shot. Single-stage detectors generally exhibit faster inference times and lower computational overhead compared to larger, two-stage detectors, and are more suited for real-time robotic applications in space~\cite{pauly_survey_2023}.    

Although the use of object detectors in spaceborne autonomy stacks is not new, current spacecraft-related works only leverage object detectors to detect objects of interest, such as uncooperative spacecraft and debris, at shorter ranges within 75 meters~\cite{pauly_survey_2023, aldahoul_localization_2022}. However, for spacecraft seeking to maneuver closer to an uncooperative spacecraft to perform OOS, the initial ranges to the target can easily exceed such distances and require long-range detection capabilities that current work do not provide. This is exacerbated by the fact that prior works only perform object detection within the visible spectrum. This means that target objects will appear small at long distances and be photometrically imperceptible due to low-light conditions in space. 

In our work, we develop long-range spacecraft detection capabilities that leverage thermal imaging which can easily highlight the spacecraft, even at long ranges, due to their extreme thermal signatures from onboard computers juxtaposed against the vacuum of space. We build upon existing work that leverage YOLO object detector variants for thermal object detection from airborne platforms~\cite{jiang_object_2022}, while also exploring new directions for spacecraft detection using zero-shot segmentation models similar to Segment Anything~\cite{kirillov2023segment}.

\subsection{Short Range Spacecraft Parts Segmentation}

Spacecraft parts identification and localization is required for OOS tasks, since they enable precise manipulation of the parts. Current methods for spacecraft parts localization operate at short range and localize parts either through object detection or segmentation. Object detection-based methods typically utilize aforementioned single-stage detectors like YOLO to predict bounding boxes encompassing particular spacecraft parts.~\cite{liu_lightweight_2022, tang_intelligent_2024, 9836634} Segmentation-based methods typically utilize popular CNN-based semantic segmentation models to perform per-pixel classification of incoming imagery. Like in object detection, large transformer-based models dominate current semantic segmentation benchmarks but are generally not suitable for real-time use. In this work, we develop a robust, real-time semantic segmentation model for spacecraft part segmentation by transferring knowledge from a large vision transformer model, Dino~\cite{caron2021emerging}, to a real-time capable FastSCNN~\cite{poudel2019fast} network.  

Aside from model selection, one notable challenge in spacecraft part localization, and field robotics in general, is the scarcity of publicly-available datasets for model training and development~\cite{lee2023online, lee2024cart, gan2023unsupervised, tang_intelligent_2024, hoang_spacecraft_2021, liu_lightweight_2022}. Some works address this issue by developing RGB spacecraft part datasets using real images, commonly scraped off the web, and synthetic images, typically rendered via Blender.~\cite{tang_intelligent_2024, hoang_spacecraft_2021} However, such works typically do not release their curated dataset to the public, which is especially common in the spacecraft industry due to the confidential nature of most projects. As such, most methods rely on techniques like transfer learning~\cite{liu_lightweight_2022} to finetune pretrained CNNs on small datasets.

In this work, we propose a method for spacecraft part segmentation that is both storage- and computationally-efficient, enabling low-latency scene perception for OOS tasks. In contrast with prior work, we use knowledge distillation to transfer learned features from a large visual-foundation model to a lightweight CNN that can run in real-time on low-compute hardware. Like other works, we develop a custom RGB dataset using our spacecraft simulator hardware~\cite{nakka2018six} for model training and benchmarking. We also validate our method on a publicly-available spacecraft parts dataset.

\subsection{Knowledge Distillation}
Knowledge distillation (KD)~\cite{hinton2015distilling, 10.1145/1150402.1150464} seeks to transfer the learned behavior of a larger model (teacher network) to a smaller, lightweight model (student network). KD is a promising technique for spacecraft applications because the deployment of models in space necessitates model compression that decreases the size of the model in order to increase inference speeds, while maintaining strong predictive performance~\cite{models_spacecraft_review}. 

In general, KD from a single teacher can be generalized into two categories: knowledge from logits and knowledge from intermediate features. When distilling knowledge from logits, the student network is trained to match the output distribution of the teacher network, usually through the cross entropy loss~\cite{hinton2015distilling, xie2018improving, 9340578}. In the latter case, knowledge is distilled by driving the features of the student to match the intermediate features of the teacher~\cite{wang2021distilling, chung2020feature, 9340578}, using some distance metric like the $L_2$ norm along with more sophisticated methods like an adversarial loss~\cite{chung2020feature}.

In our work, we utilize a simple KD method that drives the penultimate features of a FastSCNN network to match that of a large pretrained vision transformer~\cite{dosovitskiy2020image} using the $L_2$ distance metric. We use KD to pretrain our FastSCNN segmentation network before finetuning on our small spacecraft part dataset.

\section{LONG RANGE SPACECRAFT DETECTION}
\label{sec:long_range_spacecraft}
In this section, we present our method that autonomously detects an uncooperative spacecraft from approximately 100 m away from an observing spacecraft equipped with a LWIR camera in real time to allow for continuous tracking of the target.
We choose a LWIR camera for its consistency across varied environmental conditions, especially lighting, and range.
While thermal cameras at longer ranges have lower spatial resolution resulting in less detail, for long range spacecraft detection higher resolution is not needed.  Additionally, we choose a LWIR camera to align with the Edge Node mission requirements.

\begin{figure*}[htbp]
    \centering
    \includegraphics[width=\linewidth]{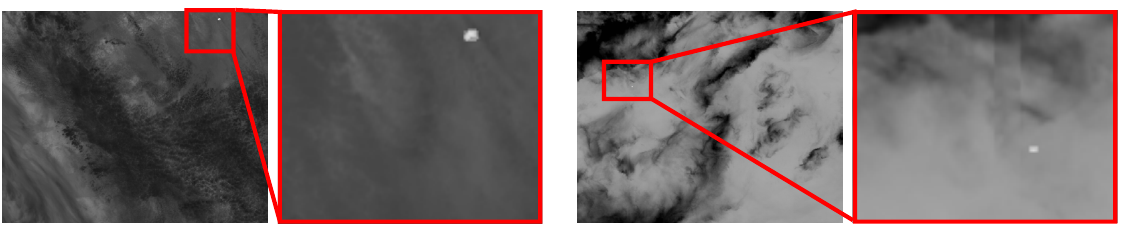}
    \captionof{figure}{Dataset samples with zoomed-in crops (red) used to develop models for long range spacecraft detection.}
    \label{fig:lwir_data_zoomed}
\end{figure*}
\subsection{LWIR Spacecraft Dataset Creation}

\subsubsection{Image Collection}
We use thermal images taken from the Landsat 8/9 satellites, which are equipped with sensors that fall within the LWIR range of the electromagnetic spectrum. These Landsat images provide the Earth background for the dataset, and vary in location, cloud coverage, and solar angle. These different features are important in order to diversify our dataset. We also use a FLIR Boson+ LWIR, model 22640A012--6IAAX, camera to capture images of the Edge Node spacecraft and remove the background.

\subsubsection{Image Processing and Data Augmentation}
We resample the spacecraft and Earth background images to align with the predicted spatial resolution properties of the mission. We assume that the ground sampling distance (GSD) is 156 m when the observing spacecraft is orbiting at 456 km.
The spacecraft additionally is assumed to be at a random distance within 20-150 m from the observing spacecraft.
The Earth background image additionally is cropped to a frame size of 100 km by 80 km. Then, the spacecraft image is superimposed on the Earth background at random positions and orientations.
At random, the spacecraft and Earth background are multiplied when superimposing to vary the contrast between spacecraft and Earth background.
We generate in total 1804 images and their annotations.
One of the key advantages of this approach is the automated labeling of images.
By controlling the placement and orientation of the spacecraft in the thermal LANDSAT imagery, we are able to generate precise labels algorithmically, eliminating the need for manual annotation.

\subsection{Spacecraft Detection using YOLOv8}
We compare a zero-shot YOLOv8 model against a finetuned domain-specific YOLOv8 model. We use YOLOv8 for our model architecture for its state-of-the-art capabilities in object detection as well as strong performance in real time applications and on edge devices~\cite{ultralytics_yolo_2023}. YOLOv8 comes in different sizes, with the smallest being YOLOv8n, which has 3.2M parameters. 
For our work, we choose the YOLOv8n-seg variant, which is pretrained on the COCO dataset~\cite{DBLP:journals/corr/LinMBHPRDZ14}. We use this baseline model to compare the finetuned zero-shot and domain specific detection model.

\subsubsection{Zero-Shot Spacecraft Detection}
Zero-shot detection is valuable in space missions as it enables the detection of uncooperative spacecraft without any prior training or knowledge of their characteristics.
To achieve zero-shot detection, the baseline YOLOv8n-seg segmentation model is further trained on a subset of the SA-1B dataset~\cite{kirillov2023segment}~\cite{mobilesamv2}, which includes segmentation masks for diverse objects. The motivation for further training on this dataset is to prevent overfitting to any specific domain. This training phase employs both bounding box and mask annotations to establish a robust detection framework. Subsequently, the model undergoes finetuning using only bounding box annotations to optimize for the specific task of spacecraft detection\footnote{This model was provided as part of MobileSAM\cite{mobilesamv2} and is available at \url{https://github.com/ChaoningZhang/MobileSAM}}.

To detect a moving, uncooperative object in space, the system incorporates onboard ephemeris data to filter out irrelevant moving background objects, such as atmospheric or celestial entities.
This filtering process utilizes optical flow methods to track detections across sequential images, applying relative velocity constraints to distinguish the target spacecraft from other moving objects.
This approach ensures the exclusion of non-target entities, thereby enhancing the precision and reliability of the spacecraft tracking system.
We assume below in validating model performance that background detections are successfully filtered out and only consider the prediction of the spacecraft in the images by picking the closest detected bounding box that has an Intersection over Union (IoU) greater than 0.5 with the ground truth.
In real application, filtering by size, lighting, or other features could be employed depending on, if any, the a priori knowledge of the target.
This zero-shot model can only detect that there is an uncooperative target and not classify the object with a label. While the model does train with classification labels, we set inference parameters to detect broadly generalizable objects. Additionally, the post-processing step to filter out background objects does not take class labels into account, ensuring the detection focuses on the presence of objects rather than their specific class labels.

\subsubsection{Domain-Specific Spacecraft Detection}
Domain-specific detection is valuable in missions where communication between spacecraft is not possible but prior knowledge of spacecraft characteristics are available. Additionally, with finetuning, we can do single stage detection and do not require multiple frames for the elimination of background detections. The single stage model learns to detect and classify the object. 
 We utilize transfer learning from the YOLOv8n-seg segmentation model trained on COCO. We finetune this pretrained model on our custom LWIR spacecraft training dataset of 1400 images and bounding box annotations. We use the default YOLOv8 training parameters, with an image size of 832 $\times$ 832, a batch size 16, and a confidence threshold of 0.01.

\subsubsection{Long Range Results}

We summarize in Table \ref{tab:LWIR_metrics} the performance of the spacecraft detection models.  To evaluate detection only in the zero-shot models, we use average precision (AP) at an IoU of 0.5 (AP50) and of 0.75 (AP75) as the metric.
As expected, the finetuned zero-shot model performs better than the baseline zero-shot model.  We use average precision (AP) calculated at various IoU thresholds for the domain specific to evaluate detection and classification.

\begin{table}[H]
\captionof{table}{Results for long-range spacecraft detection using LWIR imagery}
\label{tab:LWIR_metrics}
    \resizebox{\columnwidth}{!}{

    \begin{tabular}{p{2.5cm} p{4cm}  c c} 
    \toprule[1pt]
    Method & Model & AP50  & AP75 \\ 
    \midrule[1pt]  
    Zero-shot & YOLOv8 (COCO, B) & 0.655 & 0.437 \\
    Zero-shot & YOLOv8 (SA-1B, FT) & 0.720 & 0.425 \\
    Domain-specific & YOLOv8 (FT) & 0.890 & 0.667 \\
    \bottomrule[1pt]
    \end{tabular}
    }
    \caption*{\footnotesize B - Baseline \qquad FT - finetuned}
\end{table}

\begin{figure*}[t]
    \centering
    \includegraphics[width=\linewidth]{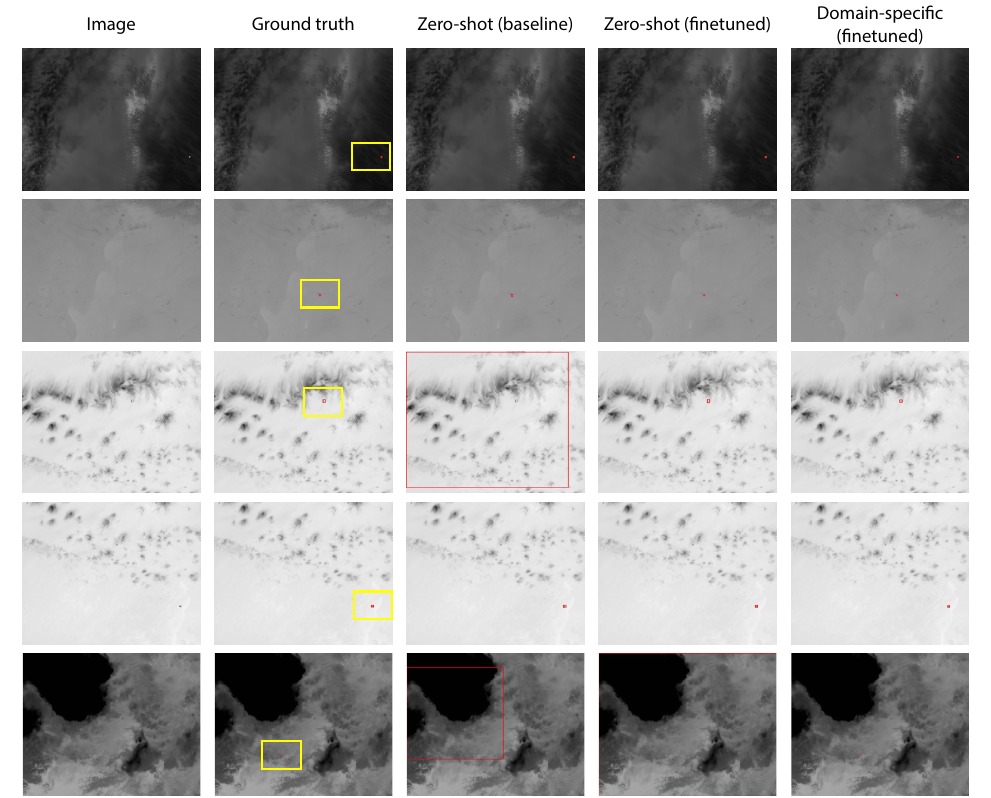}
    \captionof{figure}{Results of long range spacecraft detection using thermal imaging. To aid visualization, the spacecraft in ground truth samples are delineated with yellow bounding boxes.}
    \label{fig:lwir_results}
\end{figure*}

For long range tracking, we test inference times to ensure real time detection is feasible.  Inference times for the zero-shot detection does not include any post processing done to filter out background detections.  We perform 500 prediction passes on RGB images of dimensions $832 \times 832$ and average the inference times.  We test on a NVIDIA Jetson TX2 NX, which is the radiation tested computer chosen for the Edge Node Lite mission.~\cite{EdgeNode}

All models run between 60-90 ms on the Jetson TX2. These inference times demonstrated on the Jetson TX2 in are low enough to support real-time tracking, making it suitable for use and testing in the upcoming Edge Node Lite mission.

\section{SHORT RANGE SPACECRAFT PARTS SEGMENTATION}
In this section, we present our method to autonomously detect spacecraft parts at a short range of 20-50 m from an observing spacecraft.
We presume that, once identified via long range tracking (Section~\ref{sec:long_range_spacecraft}), the spacecraft has the capability to follow and track the uncooperative target.

For this scenario, we assume the spacecraft is equipped with a high resolution RGB camera, and switches use from the thermal camera. This change potentially sacrifices reliability under variable lighting conditions, a factor critical in space environments where lighting can be highly dynamic.  However, we gain the RGB cameras' ability to capture detailed visual information in three color channels, enhancing image detail and feature recognition.

Subsequently, the collected frames are downlinked to Earth for annotating the segmentation masks and classifications of different parts of the spacecraft, such as solar panels, body, and antenna.
Once annotated, these annotations can be uplinked back to the spacecraft\footnote{a requirement of the Edge Node mission is to do training on-board}, where they are utilized to train onboard machine learning models.
Alternatively, training can also be done on the ground with weights uploaded back to the spacecraft. Models trained on the ground, however, are subject to bandwidth uploading constraints. 

Additionally, we leverage lightweight models in order to enable future possibilities of onboard training with the goal of one day using unsupervised or self-supervised models for fully autonomous learning onboard. The intention is to enable these models to autonomously segment and classify spacecraft components in real-time using minimal images to limit the reliance on ground operations.

\subsection{Spacecraft Parts Dataset Creation}\label{sec:dataset_short_range}

\subsubsection{Custom RGB Spacecraft Parts Dataset}\label{sec:custom_dataset}
To create a semantic segmentation dataset for spacecraft parts at a shorter range, we took RGB photos of the Edge Node spacecraft in our spacecraft robotic simulator~\cite{nakka2018six} at different orientations and distances.
To streamline the labeling process, we use AnyLabeling, an AI-assisted data labeling program that uses SAM with key point prompts to generate the segmentation mask of an object.~\cite{Nguyen_AnyLabeling_-_Effortless}
We manually review the generated labels and perform minor corrections.
Our final dataset consists of 301 annotated images.
Examples of such images from the dataset can be seen in the top row of~\Cref{fig:parts_dataset}.
For training, we split the dataset into train, validation, and test sets at a 75-20-5 ratio.

\subsubsection{Adelaide Spacecraft Parts Dataset}
\label{sec:open_dataset}

\begin{figure*}[htbp]
\centering
\includegraphics[width=\textwidth]{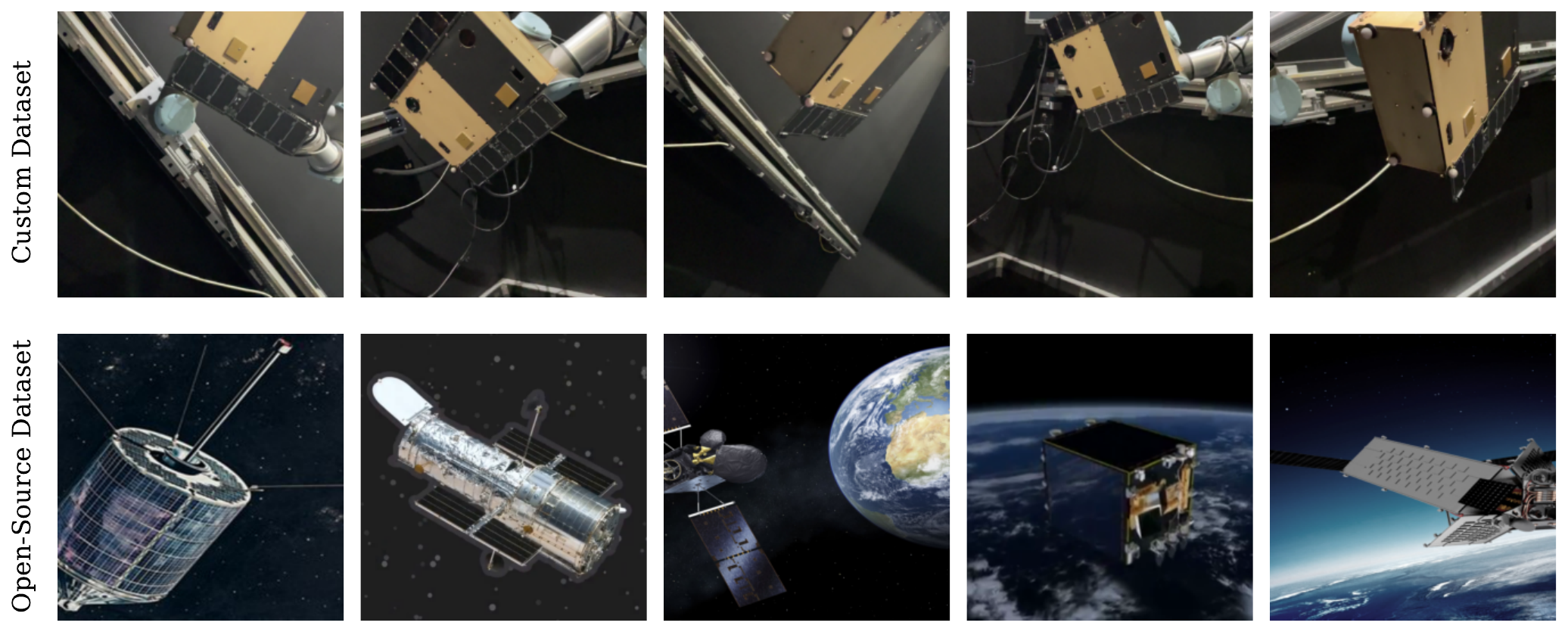}
\caption{Example images from the spacecraft parts dataset generated in the spacecraft robotic simulator (top row) and the Adelaide dataset (bottom row).}
\label{fig:parts_dataset}
\end{figure*}

To evaluate the generalizability of the segmentation models, we utilized a publicly-available dataset~\cite{hoang_spacecraft_2021} consisting of 3117 images of various satellites and space stations.
This dataset includes both synthetic and real images and videos, providing a diverse range of spacecraft types.
Unlike our custom dataset, which contains multiple images of the same spacecraft captured in a controlled lab environment, the Adelaide dataset features a wide array of spacecraft, often with only one or a few images per spacecraft.
Additionally, the backgrounds in the Adelaide dataset are more realistic, depicting space or Earth, as opposed to the controlled lab settings in our images.~\cite{hoang_spacecraft_2021}
Examples of such images from the dataset can be seen in the bottom row of~\Cref{fig:parts_dataset}.
As before, we also split this dataset into train, validation, and test sets at a 75-20-5 ratio.

\algblock{Input}{EndInput}
\algnotext{EndInput}
\algblock{Output}{EndOutput}
\algnotext{EndOutput}
\begin{algorithm}[H]
    \caption{VFM distillation into a real-time CNN}
    \label{alg:training-v2}
    \begin{algorithmic}[1]
        \State \textbf{Input:}  Networks $\mathcal{F}_\textrm{vfm}$, $\mathcal{F}_\textrm{cnn}$, training dataset $\mathcal{D}$
            \\ \qquad\quad Number of epochs N, learning rate $\eta$
        \State \textbf{Output:} Distilled CNN weights $\vb*{\theta}_\textrm{cnn}$ 
        \\
        \For{$n=1:N$}
            \State Sample image batch $\vb*{x_n}$ from $\mathcal{D}$
            \\
            \State $\vb*{z}_{\textrm{vfm}} := \mathcal{F}_\textrm{vfm}(\vb*{x}_n ; \ \vb*{\theta}_\textrm{vfm}),$ \Comment{Get VFM features} \label{alg:z_vfm}
            \State $\vb*{z}_{\textrm{cnn}} := \mathcal{F}_\textrm{cnn}(\vb*{x}_n ; \ \vb*{\theta}_\textrm{cnn}).$ \Comment{Get CNN features} \label{alg:z_cnn}
            \\
            \State $\mathcal{L} := \Vert \vb*{z}_{\textrm{cnn}} - f_\textrm{bicubic}(\vb*{z}_{\textrm{vfm}}) \Vert_2^2.$ \label{alg:loss}
            \State $\vb*{\theta}_\textrm{cnn} := \vb* {\theta}_\textrm{cnn} - \eta  \nabla_{\vb*{\theta}_\textrm{cnn}} \mathcal{L}.$\label{alg:sgd}
        \EndFor
    \end{algorithmic}
\end{algorithm}

\subsection{Spacecraft Parts Segmentation}
We outline our proposed semantic segmentation method for spacecraft parts before describing the baseline methods we compare against in our experiments. 

\subsubsection{Distilled Fast-SCNN}

Our proposed method for spacecraft segmentation utilizes the Fast-SCNN~\cite{poudel2019fast} convolutional neural network due to its low compute overhead and fast inference times on edge-compute devices~\cite{lee2024cart}.
Because real-world spacecraft datasets are limited in size and quantity, we take an additional step to pretrain the Fast-SCNN in order to prevent overfitting.
Specifically, we do this by distilling the features of a VFM (in this case, we choose Dino~\cite{caron2021emerging}) into our Fast-SCNN segmentation network features by performing student-teacher training using the ImageNet-21k dataset~\cite{5206848}, which contains approximately 14 million images of different objects, including space-related ones.
After the distillation process is complete, we finetune the Fast-SCNN segmentation network on the spacecraft dataset (\Cref{sec:dataset_short_range}).

The knowledge distillation process (\Cref{alg:training-v2}) is as follows. Given an input image $\mathcal{I}\in \mathbb{R}^{H\times W \times 3}$, we first perform a forward pass using Dino\footnote{Dino~\cite{caron2021emerging} is based on a vision transformer architecture, which operates on a sequence of $16\times16$ or $8 \times 8$ patches from the input image.}, our teacher network, and extract an intermediate feature map $\vb*{z}_{\textrm{vfm}} \in \mathbb{R}^{384 \times \floor{\frac{HW}{16\cdot 16}}}$ before reshaping it such that
\begin{equation}
    \centering
    \vb*{z}_{\textrm{vfm}} \in \mathbb{R}^{\floor*{\frac{H}{16}} \times \floor*{\frac{W}{16}} \times 384}
\end{equation}
where $\floor*{\cdot}$ denotes the floor operator. 

We then extract the intermediate feature map from the student network, Fast-SCNN\footnote{For Fast-SCNN, we update the relevant convolution kernel to have an output channel dimension of 384 in order to produce feature maps that match the channel dimension of Dino's corresponding feature map.}, by performing another forward pass to get 
\begin{equation}
    \centering
    \vb*{z}_{\textrm{cnn}} \in \mathbb{R}^{\floor*{\frac H 8} \times \floor*{\frac W 8} \times 384}
\end{equation}

To train the Fast-SCNN, we regress its feature map to that of Dino using the mean squared error loss function, updating only the weights of the Fast-SCNN via stochastic gradient descent. Note that we only backpropagate with respect to Fast-SCNN's encoder weights, since the segmentation head is not involved in the computation of the loss. During computation of the loss, we also bilinearly upsample the feature maps such that they have the same shape.

Lastly, we train the Dino-distilled Fast-SCNN on the segmentation task.
The training parameters (momentum, optimizer, learning rate) are kept consistent with the initial Fast-SCNN architecture~\cite{poudel2019fast}.

\begin{table}[H]
\centering
\captionof{table}{Short-range spacecraft part segmentation model performance (mIoU) on our custom dataset.}
\label{tab:SR_metrics_in}
\resizebox{\columnwidth}{!}{
\begin{tabular}{p{4cm}cccc} 
\toprule[1pt]
Model & All & Body & Solar Panel & Background \\ 
\midrule[1pt]
YOLOv8 & 0.967 & 0.960 & 0.950 & 0.990 \\
Fast-SCNN &  0.971 & 0.968 & 0.958 &  0.986 \\
 Dino + Fast-SCNN (ours) & 0.972  & 0.969 & 0.961 &  0.987 \\
\bottomrule[1pt]
\end{tabular}
}
\end{table}

\subsubsection{Baselines}

\paragraph{YOLOv8}
YOLOv8 is a multitask CNN capable of performing both object detection and semantic segmentation.
To create a YOLOv8 baseline for spacecraft part segmentation, we finetune the YOLOv8n-seg model variant on our custom spacecraft parts dataset, presented in Sec.~\ref{sec:dataset_short_range}, by starting from publicly-available pretrained weights to facilitate transfer learning to our space domain.
We used the default YOLOv8 parameters with an image size of $832\times832$ and trained with a batch size of 16.

\paragraph{Fast-SCNN}\label{sec:fast_scnn}

Fast Segmentation Convolutional Neural Network (Fast-SCNN) is a real-time semantic segmentation model built on high resolution image data (i.e., 1024 x 2048 px)~\cite{poudel2019fast}.
This network is designed for low power devices.
The main components of the network are a learning-to-downsample module, a coarse global feature extractor, a feature fusion module, and a standard classifier.
The global feature extractor module captures the global context for image segmentation. Then, the feature fusion merges these features using addition.

\subsection{Short Range Results}
We report the results of our short-range spacecraft part segmentation algorithms. Specifically, we report the mean Intersection over Union (mIoU) on our test set. The mIoU metric is defined as the overlap between the predicted segmentation and the ground truth mask, normalized by the total area covered by the union of the two. As such, an mIoU score of 1.0 indicates a perfect prediction, while a score of 0.0 indicates a complete mismatch.

\begin{figure}[H]
    \centering
    \includegraphics[width=\linewidth]{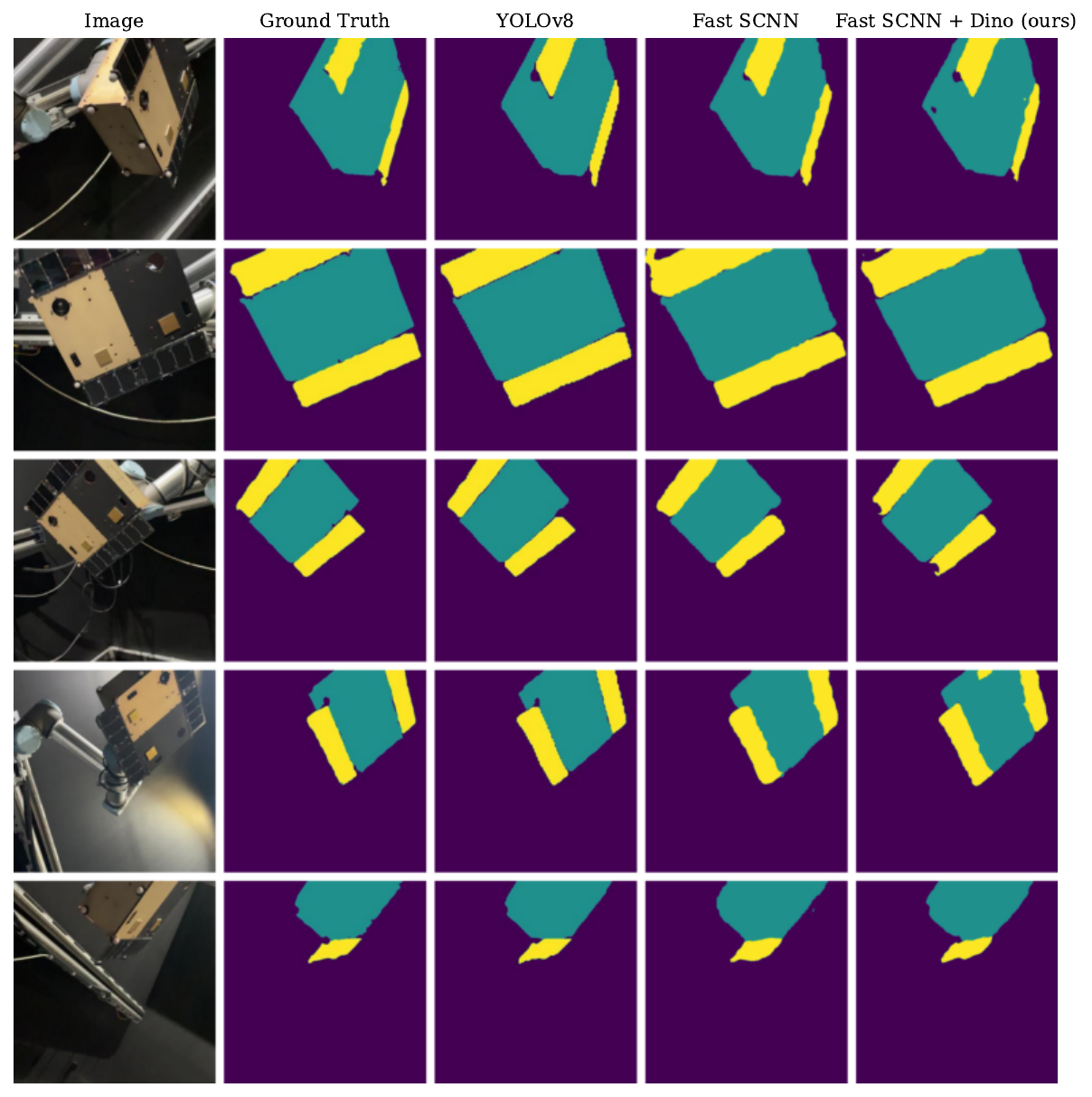}
    \captionof{figure}{Results of short range spacecraft component detection using RGB imaging.}
    \label{fig:short_range_results}
\end{figure}

We train all models on an NVIDIA Titan RTX GPU. We train for a minimum of 100 epochs and select the best performing epoch based on the mIoU of the validation dataset. We use the Adam optimizer with a learning rate of \texttt{1e-3} and a batch size of 32.
The results for training the models on the 75\%-20\%-5\% train, test, validation dataset splits are presented in~\Cref{tab:SR_metrics_in}, while in~\Cref{fig:short_range_results}, we present the qualitative results from this dataset.

\subsubsection{Spacecraft Custom Dataset Results}

On our custom spacecraft part segmentation dataset, we see that all models exhibit similar performance and nearly reach 1.0 (\Cref{tab:SR_metrics_in}). Interestingly, we found no benefit of model distillation in this dataset setting. This observation as well as the general high level of performance likely results from the smaller domain of our custom dataset. %

\subsubsection{Adelaide Dataset Results}
\begin{figure}[H]
    \centering
    \includegraphics[width=\linewidth]{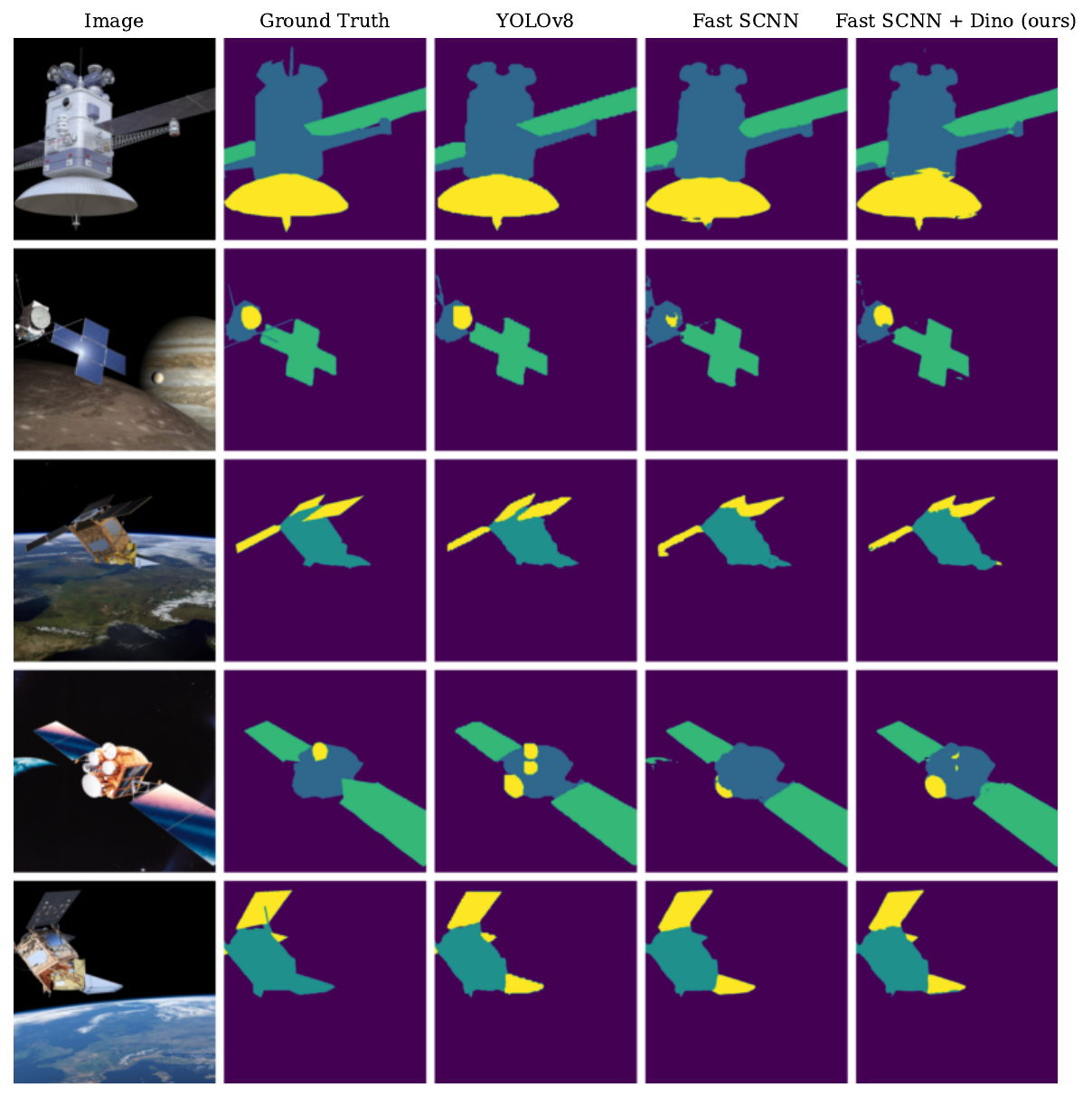}
    \captionof{figure}{Results of short range spacecraft component detection using RGB imaging on the Adelaide dataset.}
    \label{fig:short_range_open_results}
\end{figure}

We evaluate our models on the Adelaide dataset in order to quantify performance on a more difficult dataset (Table~\ref{tab:opensource_metrics_in}). In this setting, YoloV8 performs the best by a margin of 0.02. We can also see the benefits of model distillation when faced with a more difficult learning task as the distilled FastSCNN model outperforms the model trained from scratch by 0.06 mIoU. In~\Cref{fig:short_range_open_results}, we present the qualitative results from this dataset.

\begin{table}[H]
\centering
\captionof{table}{Short-range spacecraft part segmentation performance (mIoU) on the Adelaide dataset.}
\label{tab:opensource_metrics_in}
\resizebox{\columnwidth}{!}{
\begin{tabular}{p{4cm}ccccc} 
\toprule[1pt]
Model & All & Body & Solar Panel & Antenna & Background \\ 
\midrule[1pt]
YOLOv8 & 0.781 & 0.766 & 0.792 & 0.586 & 0.980\\
Fast-SCNN &0.683 & 0.705 & 0.707 & 0.347 & 0.974 \\
 Dino + Fast-SCNN (ours) & 0.756 & 0.764 &  0.768 & 0.480 & 0.982 \\
\bottomrule[1pt]
\end{tabular}
}
\end{table}

\subsubsection{Model Distillation Ablation Study}
In order to quantify the effect of our model distillation efforts for Fast-SCNN, we examine how segmentation performance is affected when fewer images are available for model training. Specifically, we examine the results when using only 75\%, 50\%, and 12.5\% of the total training set, with 5\% of the total data as validation set and 20\% as testing set.
\begin{table}[H]
\centering
\caption{Effect of training set size (\% of overall training set) on short-range spacecraft part segmentation (mIoU) for the custom dataset.
}
\label{tab:SR_split}
\resizebox{\columnwidth}{!}{
\begin{tabular}{p{4.5cm}  c c c} 
\toprule[1pt]
\multirow{2}{*}{Model} & \multicolumn{3}{c}{\% of Training Data Used} \\ \cmidrule(lr){2-4} 
 & 75\% & 50\% & 12\% \\ 
\midrule[1pt]
YOLOv8 & 0.967 & 0.967 & 0.958\\
Fast-SCNN & 0.971 & 0.966&  0.933 \\
Dino + Fast-SCNN (ours) & 0.972 &  0.969 & 0.947 \\
\bottomrule[1pt]
\end{tabular}}
\end{table}
We first perform this ablation study on our custom spacecraft parts dataset, as well as on the Adelaide dataset. \Cref{tab:SR_split} shows that the Fast-SCNN and Dino + Fast-SCNN model perform better with larger training datasets while the YOLOv8 model performs consistently well despite the size of the training dataset. 
We hypothesize that the consistent performance of YOLOv8 is due to its extensive use of data augmentation compared to our Fast-SCNN variants.

\begin{table}[H]
\centering
\caption{Effect of training set size (\% of overall training set) on short-range spacecraft part segmentation (mIoU) for the Adelaide dataset. }
\label{tab:opensource_split}
\resizebox{\columnwidth}{!}{
\begin{tabular}{p{4cm} c c c} 
\toprule[1pt]
\multirow{2}{*}{Model} & \multicolumn{3}{c}{\% of Training Data Used} \\ \cmidrule(lr){2-4} 
 & 75\% & 50\% & 12\% \\ 
\midrule[1pt]
YOLOv8 & 0.781 & 0.709 & 0.635 \\
Fast-SCNN &  0.683 & 0.598 & 0.537\\
Dino + Fast-SCNN (ours) & 0.756 & 0.722 & 0.636\\
\bottomrule[1pt]
\end{tabular}
}
\end{table}

\begin{figure}[H]
    \centering
    \includegraphics[width=\columnwidth]{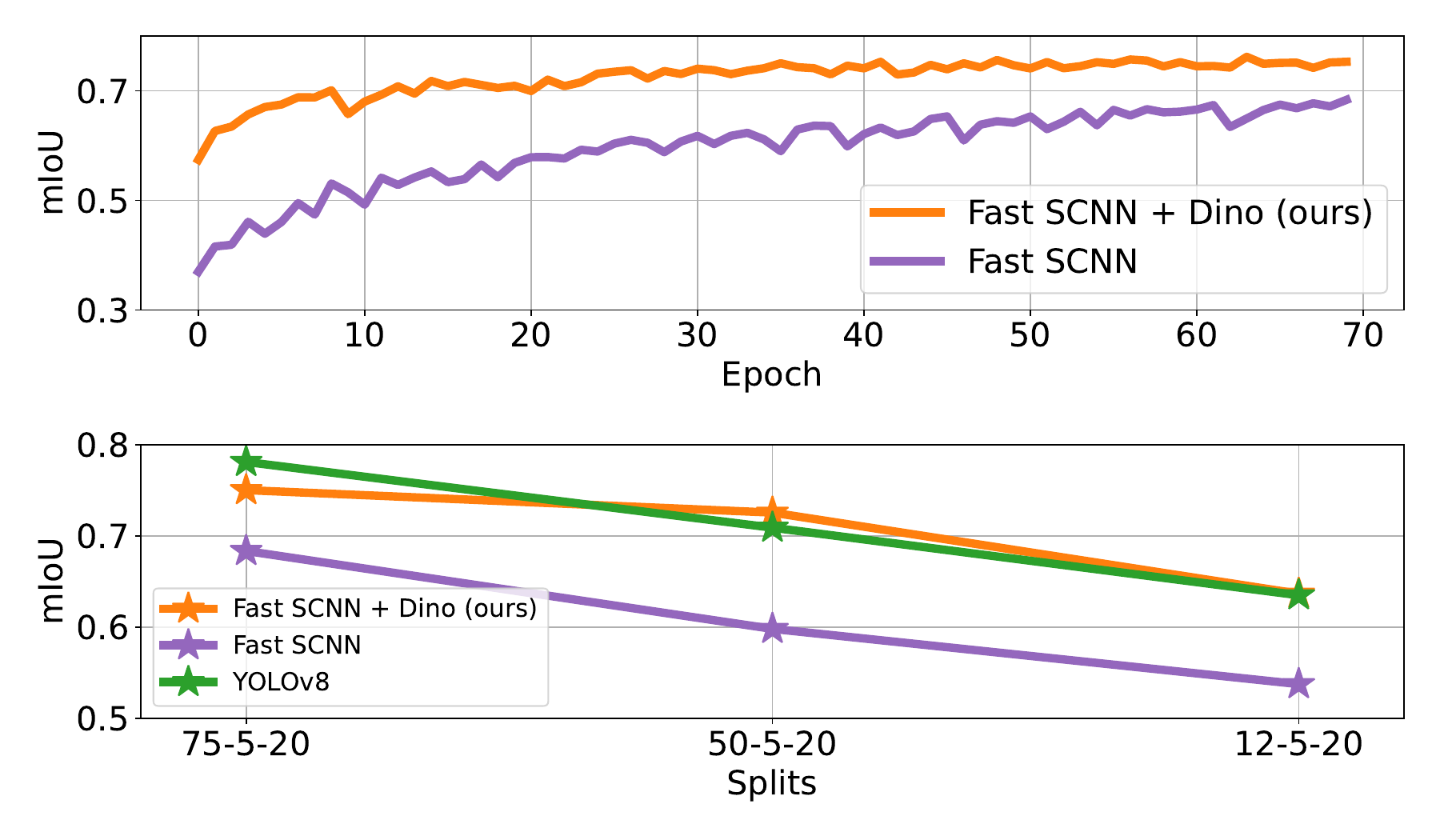}
    \caption{(Top) mIoU comparison between Fast-SCNN and our method (Fast-SCNN + Dino) for the 75-20-5 splits. We show that pretraining with Dino features improves performance. (Bottom) Best mIoU performance for different splits of the training dataset.}\label{fig:performance_dino_plots}
\end{figure}

In addition to our own dataset, we perform the same study on a more difficult dataset (\Cref{sec:open_dataset}) with added class complexity.~\cite{hoang_spacecraft_2021} 
We show these results in~\Cref{tab:opensource_split} and~\Cref{fig:performance_dino_plots}.
Because of the higher complexity of the dataset, all models performance tends to decrease with the decrease of the training dataset.
In addition, our results (top plot of \Cref{fig:performance_dino_plots}) on this dataset show that distillation noticeably increases training convergence speed and overall model performance, which could be a useful behavior for compute-limited onboard training.
As such, the more complicated a dataset becomes, the more we can see benefits of distillation versus training from scratch.

\begin{figure}[H]
    \centering
    \includegraphics[width=\columnwidth]{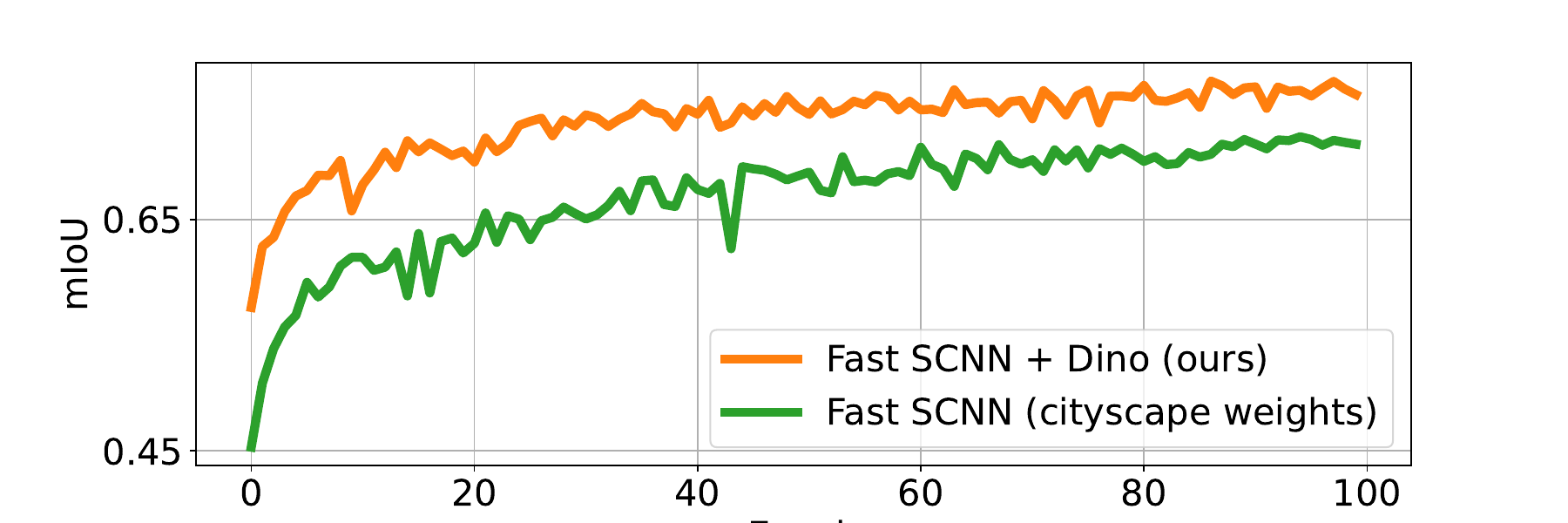}
    \caption{Comparison between our method and pre-training Fast-SCNN with other available weights.}
\label{fig:comparison_dino_citys}
\end{figure}

In~\Cref{fig:comparison_dino_citys}, we investigate whether KD with Dino features increases the performance, compared to only using pre-trained weights (i.e., Cityscapes weights~\cite{Cordts2016Cityscapes}).
The results show that employing KD with Dino slightly increases the overall mIoU performance.

\begin{table}[H]
\centering
\captionof{table}{Compute Results.}
\label{tab:hardware_metrics}
\resizebox{\linewidth}{!}{
\begin{tabular}{p{4cm} p{2.5cm} ccc}
\toprule[1pt]
\multirow{2}{*}{Model} & \multirow{2}{*}{Task} & \multicolumn{2}{c}{Inference Time (ms)} & \multirow{2}{*}{\makecell{Params \\ (M)}} \\ \cmidrule(lr){3-4} 
                   & & TX2 & Titan RTX &  \\ 
\midrule[1pt]
YOLOv8 & Instance Segm. & 78.70 & 6.41 & 3.40 \\
YOLOv8 & Semantic Segm.$^\dagger$ & 363.02 & 31.29 & 3.40 \\
Fast-SCNN & Semantic Segm. & 49.39  &  3.54 & 1.13 \\
Dino + Fast-SCNN (ours) & Semantic Segm. & 92.56 & 4.29 & 1.55\\
\bottomrule[1pt]
\end{tabular}
}
\captionof*{table}{\footnotesize $^\dagger$ YOLOv8 outputs binary masks per detected instance and requires an additional postprocessing step to create semantic segmentation masks.}

\end{table}

\subsubsection{Computational Benchmarks}

We benchmark the inference time of the short range detection models in Table~\ref{tab:hardware_metrics}.
We report results on the Nvidia Jetson TX2 (256 CUDA cores) since its size and power usage is ideal for low-compute robotic operations. We also provide benchmarks on an NVIDIA Titan RTX (4608 CUDA cores) for comparison to a workstation GPU.

Overall, all models, besides Yolov8 (semantic segmentation), can provide predictions at a 10 Hz minimum (\Cref{tab:hardware_metrics}). We find that these results are suitable for real-time use onboard the Edge Node Lite mission.
Fast-SCNN variants notably require less than half the storage requirements compared to YoloV8. While storage is less of a concern in terrestrial field robotics, it is vital to keep in check for when developing perception algorithms for spacecraft due to the limited bandwidth if performing model updates from Earth to space.

\section{CONCLUSION}
We present a long range detection methodology of uncooperative targets in space utilizing zero-shot detection and domain specific, single-shot detection with YOLOv8. Additionally, we present a short range detection methodology to segment the components of the uncooperative spacecraft potentially utilizing onboard training. We propose using long range detection to identify uncooperative targets, such as space debris and inactive satellites, and segmentation at shorter ranges to learn more detailed features of the targets to increase space situational awareness. 

For the short range, we proposed a method for spacecraft part segmentation that
leverages visual foundation models distilled into a lightweight fast semantic segmentation network (Fast-SCNN) to increase segmentation performance while keeping inference time and onboard storage requirements low to meet potential onboard training requirements. 
An additional benefit to distilling the Fast-SCNN from a visual foundation model such as Dino is less storage overhead, making it easier to deploy and update on an edge device in space compared to the larger YOLOv8.
We plan to test the long range detection and onboard training and segmentation on the upcoming Aerospace Corporation Edge Node Lite mission in 2025.~\cite{EdgeNode}

\subsection*{Acknowledgments}
This work was funded by The Aerospace Corporation. 
We thank S. Kishore and C. Smith for their work in validating our algorithm on-board the Jetson TX2 in the spacecraft robotic simulator; J. Preiss for his suggestions related to the open-source dataset and the comparison plot; S. Binkin for his work on adding textures on the EdgeNode spacecraft in Blender; F. Nasriddinov for his initial work in validating Segment Anything on the thermal spacecraft dataset; J. Cho and M. Anderson for their help with docker and ROS for the NVIDIA Jetson TX2.

\bibliography{bib.bib}
\bibliographystyle{unsrt}

\end{multicols*}
\end{document}